\documentclass[conference]{IEEEtran}
\IEEEoverridecommandlockouts
\usepackage{cite}
\usepackage{amsmath,amssymb,amsfonts}
\usepackage{algorithmic}
\usepackage{graphicx}
\usepackage{textcomp}
\usepackage{xcolor}

\usepackage{multirow}
\usepackage{booktabs}
\usepackage[ruled,vlined]{algorithm2e}
\usepackage{subcaption}   

\def\BibTeX{{\rm B\kern-.05em{\sc i\kern-.025em b}\kern-.08em
    T\kern-.1667em\lower.7ex\hbox{E}\kern-.125emX}}
\begin{document}

\title{Hardware-Aware DNN Compression for Homogeneous Edge Devices}

\author{
\textbf{Kunlong Zhang$^1$\;~Guiying Li$^{2,*}$~\;~Ning Lu$^{1}$~\;~Peng Yang$^1$~\;~Ke Tang$^1$} \\
$^1$Department of Computer Science and Engineering, \\ Southern University of Science and Technology, Shenzhen, China \\
$^2$Pengcheng Laboratory, Shenzhen, China \\
{\centering $^*$Corresponding author. Email: ligy@pcl.ac.cn}
}



\maketitle

\begin{abstract}

Deploying deep neural networks (DNNs) across homogeneous edge devices (the devices with the same SKU labeled by the manufacturer) often assumes identical performance among them. However, once a device model is widely deployed, the performance of each device becomes different after a period of running. This is caused by the differences in user configurations, environmental conditions, manufacturing variances, battery degradation, etc. Existing DNN compression methods have not taken this scenario into consideration and can not guarantee good compression results in all homogeneous edge devices. 
To address this, we propose Homogeneous-Device Aware Pruning (HDAP), a hardware-aware DNN compression framework explicitly designed for homogeneous edge devices, aiming to achieve optimal average performance of the compressed model across all devices. To deal with the difficulty of time-consuming hardware-aware evaluations for thousands or millions of homogeneous edge devices, HDAP partitions all the devices into several device clusters, which can dramatically reduce the number of devices to evaluate and use the surrogate-based evaluation instead of hardware evaluation in real-time. 
Extensive experiments on multiple device types (Jetson Xavier NX and Jetson Nano) and task types (image classification with ResNet50, MobileNetV1, ResNet56, VGG16; object detection with YOLOv8n) demonstrate that HDAP consistently achieves lower average latency and competitive accuracy compared to state-of-the-art methods, with significant speedups (e.g., 2.86$\times$ on ResNet50 at 1.0G FLOPs). HDAP offers an effective solution for scalable, high-performance DNN deployment methods for homogeneous edge devices.

\end{abstract}



\vspace{-0.3cm}

\section{Introduction}
\label{sec:intro}
Recently, the development of Artificial Intelligence \& Internet of Things (AIoT) has raised an increasing demand for deploying AI models on edge devices~\cite{AIoT1}. 
However, as models become more powerful, their size, computational requirements, and latency grow proportionally. Modern neural networks, with millions or even billions of parameters (e.g., GPT-3~\cite{gpt3}), encounter increasing challenges in widespread deployment, especially due to stringent energy and latency constraints~\cite{AIoT1}.
In scenarios such as autonomous driving, failing to meet latency constraints not only diminishes user experience but also raises serious safety risks.
In practice, such DNNs should be carefully compressed for edge deployment, and many researchers are continually working on the issue~\cite{
Structured_Pruning_Survey
}.

Hardware-aware DNN compression~\cite{HALP,Multi-Device,HAMP,AMC,ECC} offers an effective solution to reduce model size for edge devices. "Hardware-aware" here indicates evaluating the compressed models on edge devices for their real performances, which is different from traditional DNN compression methods where proxy metrics (like FLOPs) are adopted as the edge performances of compressed models. The actual feedback from edge devices makes hardware-aware DNN compression the best suitable for AIoT scenarios.

\begin{figure}[t!]
    \centering
    \begin{subfigure}[t]{0.23\textwidth}
        \centering
        \includegraphics[width=\linewidth]{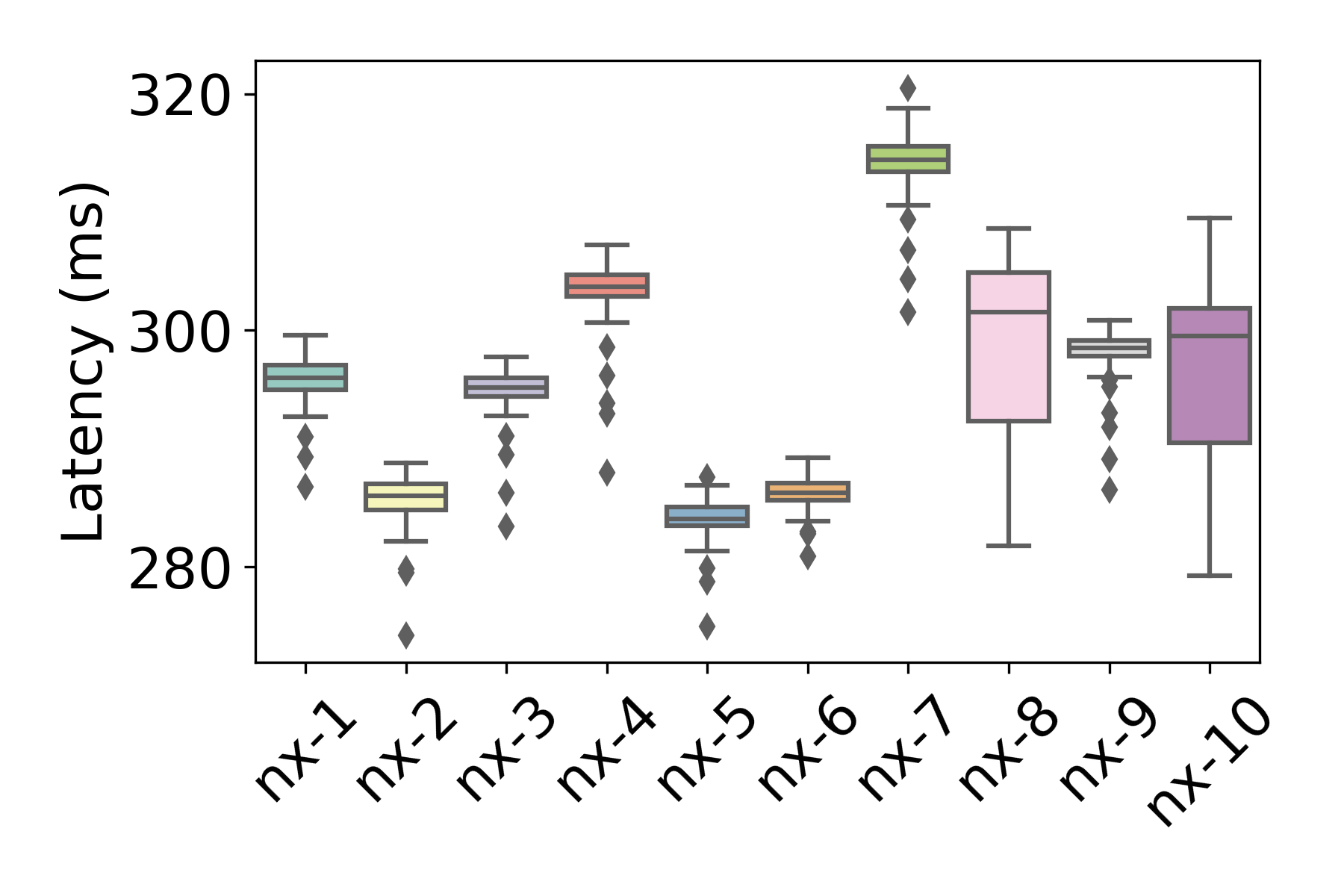}
        \vspace{-0.8cm}
        \caption{}
        \label{fig:nx_box}
    \end{subfigure}
    \hfill
    \begin{subfigure}[t]{0.23\textwidth}
        \centering
        \includegraphics[width=\linewidth]{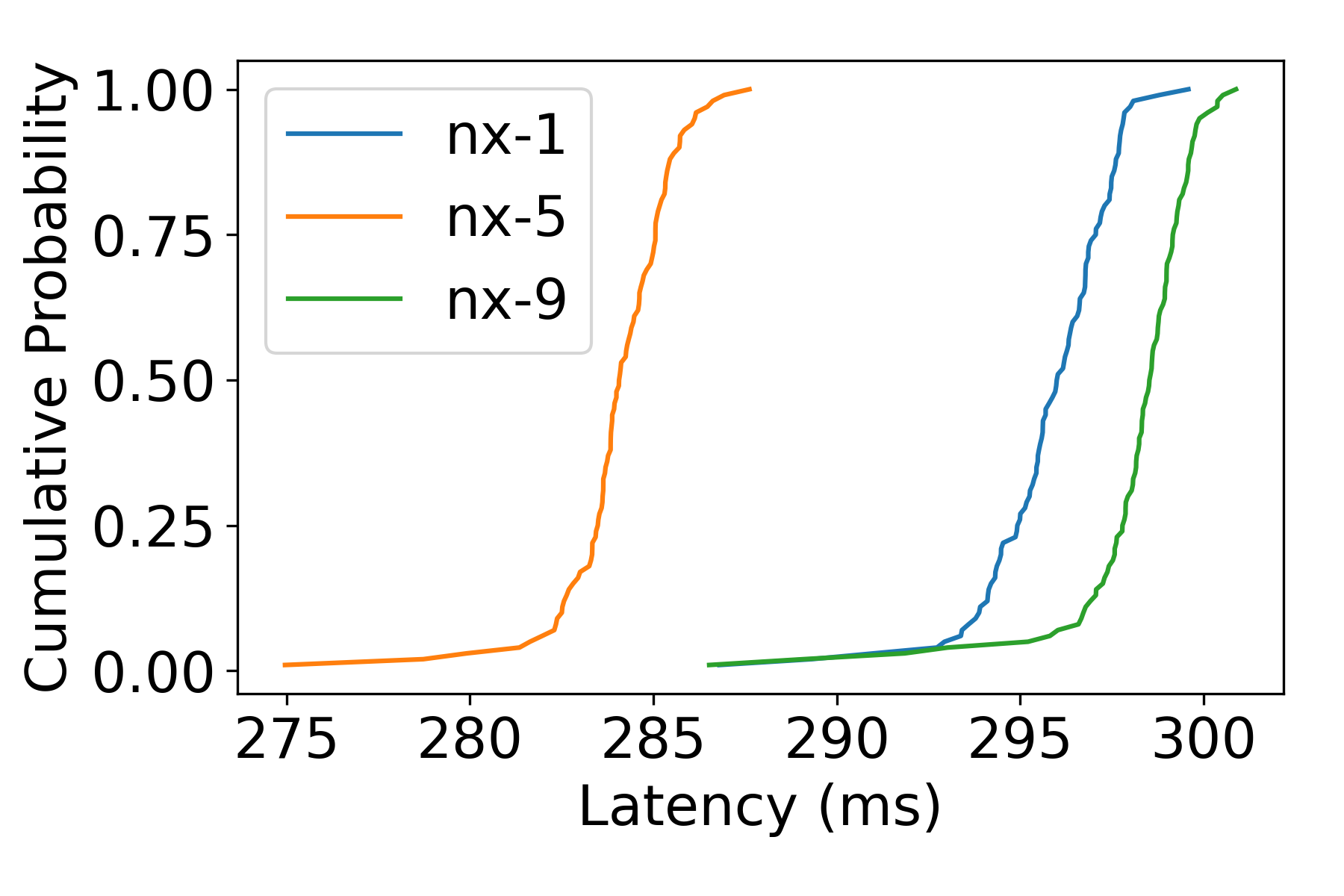}
        \vspace{-0.8cm}
        \caption{}
        \label{fig:nx_cdf}
    \end{subfigure}
    \hfill
    \begin{subfigure}[t]{0.23\textwidth}
        \centering
        \includegraphics[width=\linewidth]{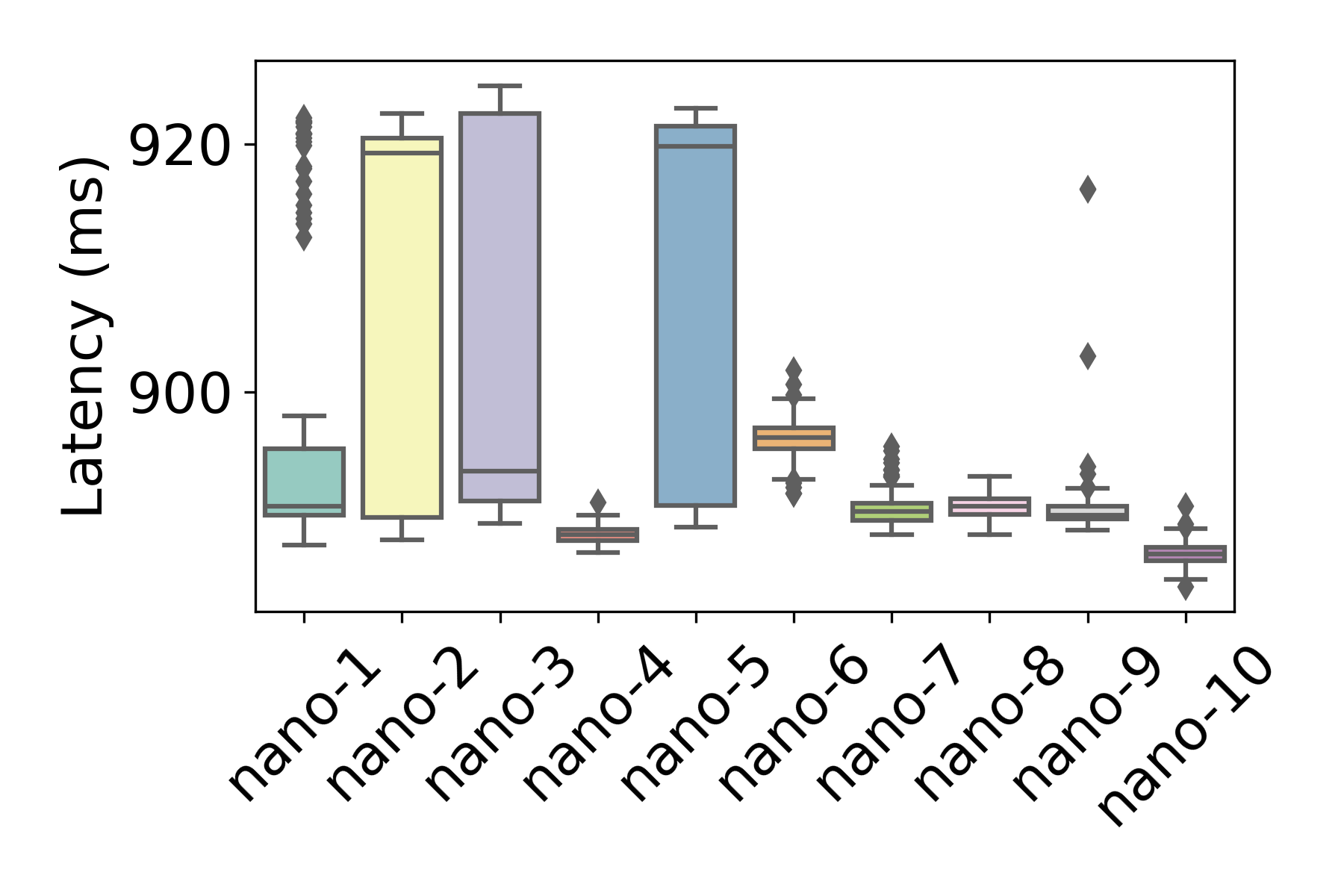}
        \vspace{-0.8cm}
        \caption{}
        \label{fig:nano_box}
    \end{subfigure}
    \hfill
    \begin{subfigure}[t]{0.23\textwidth}
        \centering
        \includegraphics[width=\linewidth]{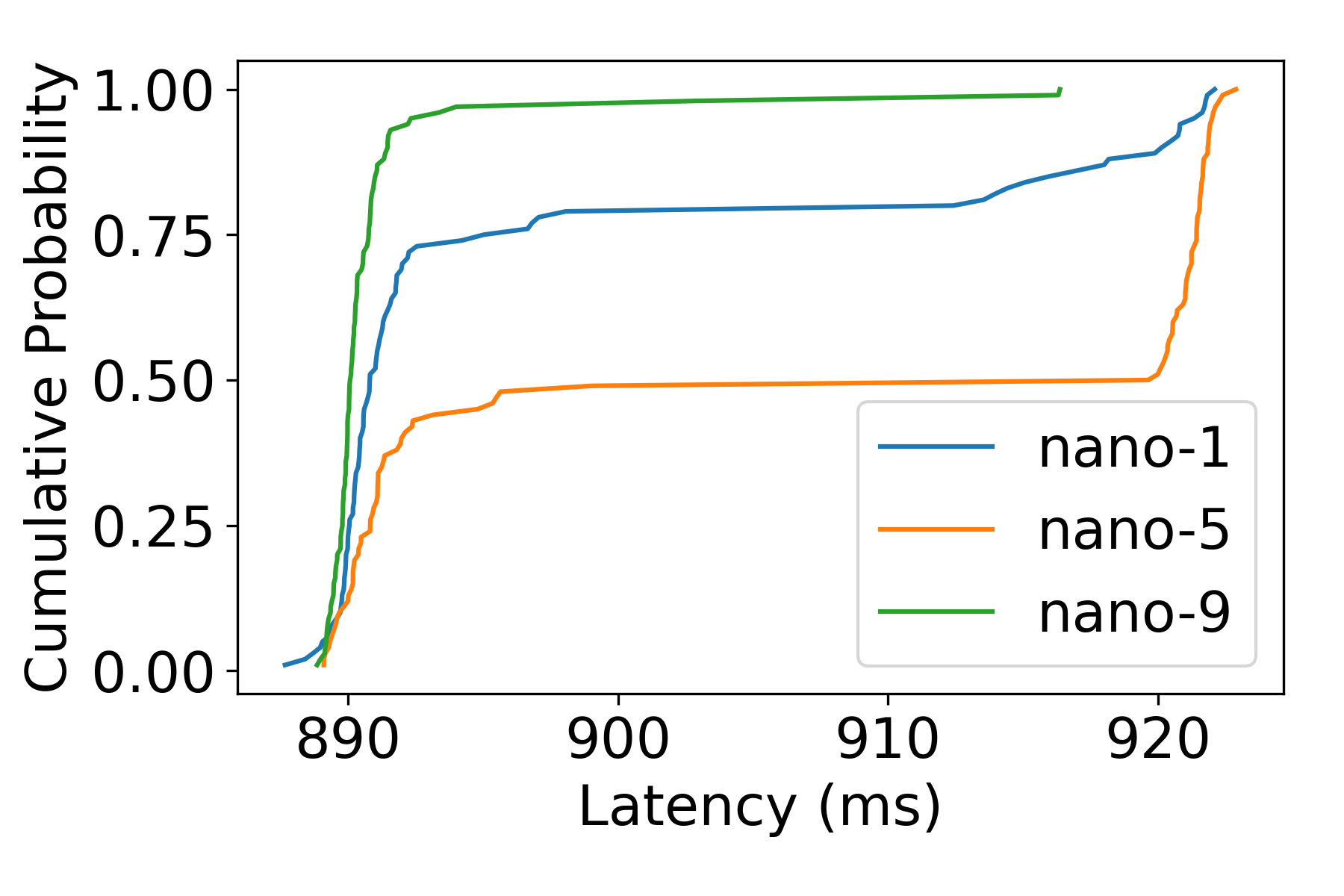}
        \vspace{-0.8cm}
        \caption{}
        \label{fig:nano_cdf}
    \end{subfigure}
    \caption{Inference latency of a DNN model on homogeneous edge clusters. (a), (c): noticeable latency variation across devices in Jetson Xavier NX (nx) and Nano clusters (nano). (b), (d): significant differences in cumulative distribution function curves of three sample devices per cluster.}
    \label{fig:device_variance}
    \vspace{-0.5cm}
    
\end{figure}

Despite the effectiveness, existing hardware-aware DNN compression methods are not yet practical for large-scale deployment.
First, most existing methods are designed for \textbf{a single edge device}~\cite{HALP,Multi-Device,HAMP,AMC,ECC}, whereas the real-world target is \textbf{a homogeneous edge cluster}—a large number of devices with identical architectures (e.g., deploying a model to millions of iPhone 16). 
However, prior studies~\cite{Performance_Variance,Performance_Variance_facebook} have revealed \textbf{non-negligible performance variation} among such devices, caused by user configurations, environmental conditions, or manufacturing inconsistencies (see Fig.~\ref{fig:device_variance}).
Such performance variation violates the fundamental assumption of existing methods—that all homogeneous devices have the same performance—thereby limiting their applicability to deployment on homogeneous device clusters.
Second, hardware-aware evaluation is costly and impractical for hundreds or thousands of homogeneous edge devices~\cite{Time_consuming}.
Each hardware-aware evaluation requires running the candidate compressed model on an edge device tens of hundreds of times to obtain an average performance.

This paper proposes a new hardware-aware DNN compression method, \textbf{H}omogeneous-\textbf{D}evice \textbf{A}ware \textbf{P}runing (\textbf{HDAP}), to address the aforementioned challenges. HDAP formulates the compression task as a constrained single-objective optimization problem that minimizes the average inference latency across homogeneous devices while preserving model accuracy.
It adopts structured pruning~\cite{Structured_Pruning_Survey} to remove redundant components from trained DNNs. The optimal compressed model is obtained by solving the proposed optimization problem.
Two key challenges arise: designing an effective optimization algorithm and reducing the cost of hardware evaluation. To this end, HDAP employs a derivative-free evolutionary algorithm, \textbf{negatively correlative search} (NCS)~\cite{NCS}, inspired by the success of evolutionary algorithms in complex optimization tasks~\cite{evolution1,evolution2,evolution3,evolution4,evolution5,evolution6}, and replaces hardware evaluation with a surrogate-based approach to accelerate the search process.

Our main contributions are summarized as follows:
\begin{itemize}
    \item We identify and formulate the overlooked problem of hardware-aware DNN compression for homogeneous edge device clusters. Existing methods assume identical performance across devices, which does not hold in practice and limits deployment effectiveness.

    \item We propose HDAP, a hardware-aware pruning framework that improves average latency and deployment consistency across homogeneous edge devices via surrogate evaluation and iterative pruning with fine-tuning.

    \item We evaluate HDAP on multiple tasks (image classification and object detection) and device types (Jetson Xavier NX and Jetson Nano), demonstrating its effectiveness in improving both compression efficiency and cross-device performance consistency.
\end{itemize}


    

\vspace{-0.1cm}

\section{Related Work}
\label{sec:related_work}
\subsection{Hardware-Aware DNN Compression}

Hardware-aware DNN compression methods aim to optimize DNN models not only for accuracy but also for performance on specific hardware devices. These methods integrate hardware performance metrics, such as inference latency~\cite{HALP}, memory usage~\cite{HAMP,AMC}, and energy~\cite{ECC}, directly into the compression or optimization process to produce compressed models that are suited for deployment on the target hardware. 
However, these methods are designed for one edge device, assuming consistent performance across homogeneous edge devices—a condition that often does not hold in practice~\cite{Performance_Variance,gpu_device}.
Although some research in hardware-aware compression addresses multi-device scenarios, it primarily focuses on different types of devices (e.g., GPUs, embedded systems, mobile phones), where “multi-device” refers to device diversity~\cite{Multi-Device}. In contrast, this work focuses on homogeneous edge devices.
\subsection{Homogeneous Edge Devices}
Homogeneous edge devices are devices with the same model, which widely exists in AIoT, like the distributed replicas of mobile phones labeled as the same SKU. Existing hardware-aware DNN compression methods always assume that those devices have exactly the same ability, which is not true in reality. Prior study~\cite{Performance_Variance,Performance_Variance_facebook} indicates that running the same DNN model on homogeneous edge devices will lead to substantial variability in performance, where power consumption can vary by 10\% to 40\%, and runtime differences ranging from 6\% to 20\%~\cite{Performance_Variance}. This means the existing hardware-aware DNN compression can get misleading feedback from the edge and can not guarantee a good compressed model. In this paper, HDAP proposes a new way to identify the differences among homogeneous devices by clustering and can effectively compress DNN model in the scenario.




\section{Methodology}
\label{sec:methodology}
In this section, we present HDAP, a method for hardware-aware DNN compression on homogeneous edge devices. HDAP formulates the compression task as a constrained optimization problem and solves it using a population-based search algorithm. To reduce evaluation overhead, it clusters devices into a few representative types. To further accelerate latency estimation, HDAP employs surrogate models in place of real-time hardware evaluation.

\vspace{-0.3cm}
\begin{figure}[ht]
    \centering
    \begin{subfigure}[t]{0.30\linewidth}
        \centering
        \includegraphics[width=\linewidth]{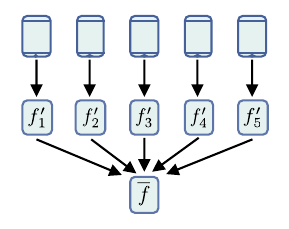}
        \caption{Per-device}
        \label{fig:surrogate_method_b}
    \end{subfigure}
    \hfill
    \begin{subfigure}[t]{0.30\linewidth}
        \centering
        \includegraphics[width=\linewidth]{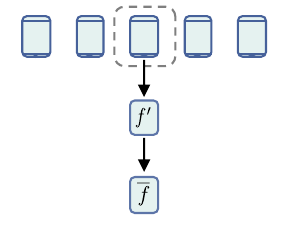}
        \caption{Unified}
        \label{fig:surrogate_method_a}
    \end{subfigure}
    \hfill
    \begin{subfigure}[t]{0.30\linewidth}
        \centering
        \includegraphics[width=\linewidth]{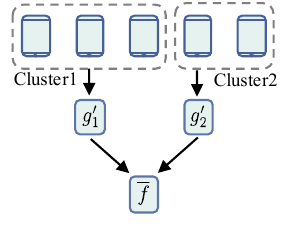}
        \caption{Clustering-based}
        \label{fig:surrogate_method_d}
    \end{subfigure}
    \caption{Different hardware evaluation methods.}
    \label{fig:surrogate_model_construction_methods}
\end{figure}
\vspace{-0.6cm}

\subsection{Problem Formulation}

\label{sec:problem_formulation}
We aim to compress a pre-trained DNN model $M$ into $M^*$ that minimizes the average inference latency across homogeneous edge devices, subject to an accuracy constraint. Let $M$ have $L+1$ layers, and define a structured pruning operator $P$ with a pruning vector $\boldsymbol{X} = [x_1, \dots, x_L]$, where each $x_l \in [0, 1)$ denotes the pruning ratio at layer $l$. The compressed model is given by $M' = P(M, \boldsymbol{X})$, where $P(\cdot)$ removes redundant filters or neurons based on their $L_2$-norm importance. Let $\mathcal{C} = \{c_1, \dots, c_N\}$ be the set of devices, and $f_i(M')$ denote the latency of model $M'$ on device $c_i$. The average latency is defined as:
\vspace{-0.15cm}
\begin{equation}
\overline{f}(M') = \frac{1}{N} \sum_{i=1}^{N} f_i(M').
\label{eq:average_inference_latency}
\end{equation}
\vspace{-0.35cm}

\begin{figure*}[ht]
    \centering
    \includegraphics[width=.83\linewidth]{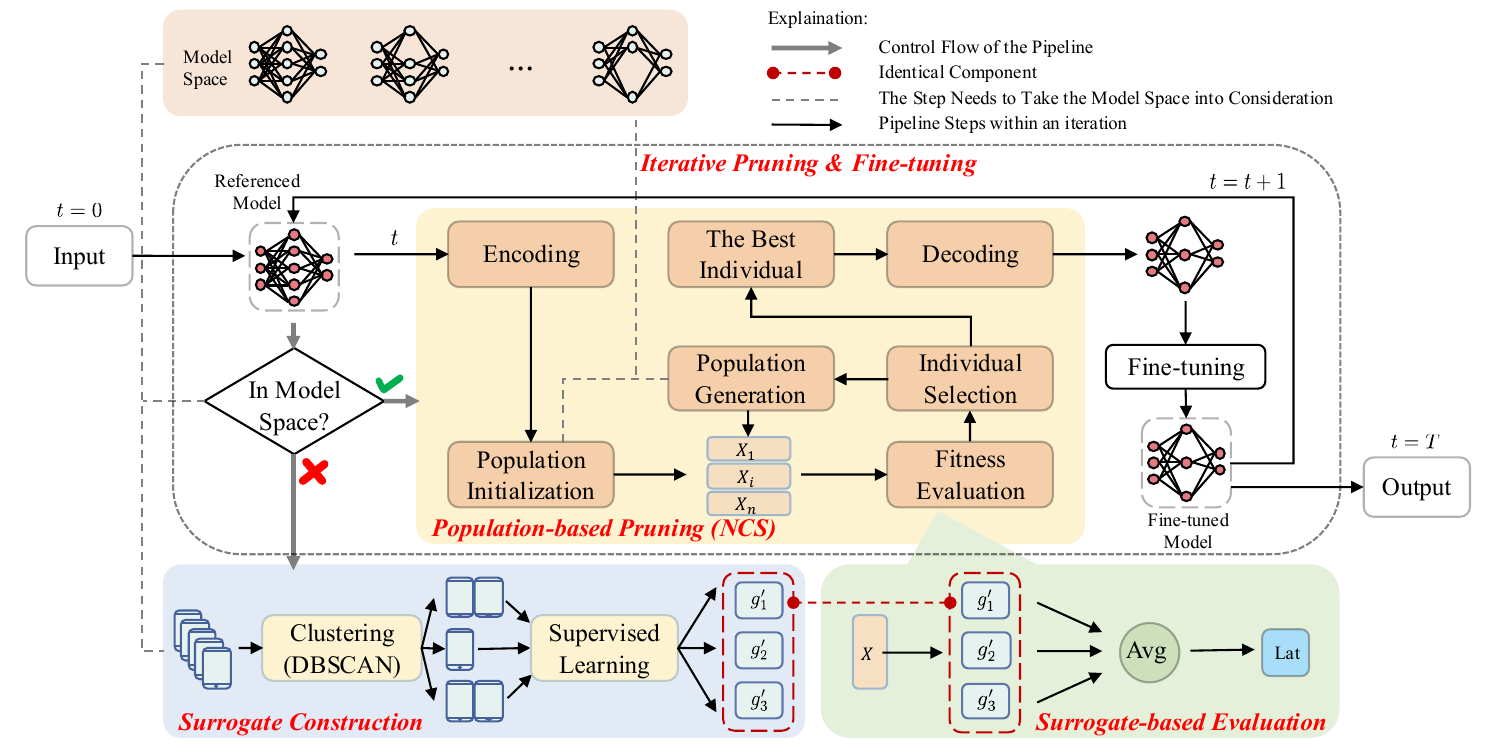}
    \vspace{-0.1cm}
    \caption{Overview of HDAP. It consists of iterative pruning guided by surrogate-based evaluation, and fine-tuning. Surrogates are built via clustering and supervised learning if $M$ is outside the model space.}

    \label{fig:pipeline}
\vspace{-0.5cm}
    
\end{figure*}

However, evaluating~\eqref{eq:average_inference_latency} is impractical when $N$ is large, as it requires running candidate models on every device. In real-world edge scenarios, unstable networks and limited computational resources make it difficult to obtain $f_i(\cdot)$ reliably and efficiently, resulting in prohibitive evaluation costs (see Fig.~\ref{fig:surrogate_method_b}). A common alternative is to use a unified latency evaluation~(Fig.~\ref{fig:surrogate_method_a}), which ignores the performance variation across homogeneous devices and introduces significant noise, making it difficult to guarantee consistent performance of the compressed model $M'$ across homogeneous devices.

To address this challenge, HDAP introduces a clustering-based approach to reduce the number of evaluations by grouping devices with similar performance into $K$ clusters, where $K \ll N$ (see Fig.~\ref{fig:surrogate_method_d}).  
Formally, HDAP partitions the device set \( \mathcal{C} \) into $K$ non-overlapping clusters \( \mathcal{C}_1, \dots, \mathcal{C}_K \), satisfying:
\vspace{-0.15cm}
\begin{equation}
    \begin{aligned}
    &\mathcal{C} = \mathcal{C}_1 \cup \dots \cup \mathcal{C}_K, \\
    &\mathcal{C}_k \cap \mathcal{C}_j = \emptyset, \quad \forall k \ne j, \\
    &|\mathcal{C}_k| > 0, \quad \forall k \in \{1, \dots, K\}.
    \end{aligned}
    \label{eq:clustering}
\end{equation}
\vspace{-0.25cm}

This clustering-based method aligns with real-world deployment. Although homogeneous edge devices share the same architecture, they exhibit different performance due to limited factors such as manufacturing variations and user-specific configurations. 
These factors remain stable over time, causing devices to naturally form clusters with similar performance, which supports HDAP's clustering design.
Furthermore, since devices within each cluster exhibit similar performance, latency evaluation can be performed on a subset of representative devices, reducing hardware evaluation overhead.
Consequently,~\eqref{eq:average_inference_latency} can be reformulated as:
\vspace{-0.15cm}
\begin{equation} 
\overline{f}(M') = \frac{1}{K} \sum_{k=1}^{K} g_k(M'),
\label{eq:average_inference_latency_cluster}
\end{equation}

{
\vspace{-0.25cm}
\setlength{\parindent}{0cm}
where $g_k(M')$ denotes the latency of $M'$ on cluster $\mathcal{C}_k$.
}

In~\eqref{eq:average_inference_latency_cluster}, evaluating candidate models on cluster $\mathcal{C}_k$ is time-consuming due to limited device capability and the large number of candidate models involved. To address this, HDAP introduces a data-driven surrogate modeling approach for homogeneous edge scenarios, where latency is predicted instead of measured on-device. The prediction task is formulated as a regression problem: given a compressed model $M'$ and device type $k$, the surrogate model $g_k'(\cdot; \theta_k)$ is trained to approximate the true latency $g_k(M')$ by minimizing:
\vspace{-0.15cm}
\begin{equation}
 \theta_k^* = \mathop{\arg\min}_{\theta_k} \sum_{M'} \left( g_k(M') - g_k'(M'; \theta_k) \right)^2.
 \label{eq:surrogate_objetive_function}
\end{equation}
\vspace{-0.3cm}

By leveraging device clustering and surrogate modeling, HDAP efficiently approximates the average inference latency without real-time hardware evaluation:
\vspace{-0.15cm}
\begin{equation} 
\overline{f}(M') \approx \frac{1}{K} \sum_{k=1}^{K} g'_k(M';\theta_k). 
\end{equation}
\vspace{-0.15cm}
Finally, the DNN compression problem is then formulated as:
\begin{equation}
\begin{aligned}
    \boldsymbol{X}^* = \mathop{\arg\min}_{M' = P(M, \boldsymbol{X})} \; & \overline{f}(M') \\
    \text{s.t.} \quad & \text{Acc}(M') \geq \alpha  \text{Acc}(M),
\end{aligned}
\label{eq:optimization_problem}
\end{equation}
where $\alpha \in (0,1]$ denotes the acceptable accuracy ratio.

\subsection{Method Overview}

To solve the optimization problem in~\eqref{eq:optimization_problem}, HDAP adopts an iterative framework comprising two core components: 1) \textbf{population-based pruning} and 2) \textbf{surrogate-based evaluation}, as shown in Fig.~\ref{fig:pipeline}.

In each iteration, HDAP generates candidate pruning vectors $\boldsymbol{X}$ via a population-based evolutionary strategy. Each individual represents a pruned model $M' = P(M, \boldsymbol{X})$, and its average latency $\overline{f}(M')$ is estimated by surrogate-based evaluation and used as the fitness value. The best pruning vector is then selected and fine-tuned to restore accuracy.

To enable efficient evaluation, HDAP constructs surrogate models $g_k(\cdot;\theta_k)$ for each device cluster. These surrogate models are learned through a two-stage process: (i) \textbf{unsupervised device clustering} using DBSCAN~\cite{DBSCAN2} based on the latency of a benchmark model, and (ii) \textbf{supervised learning for surrogates}, where surrogate models are trained to predict the latency of pruned models, using pruning vectors $\boldsymbol{X}$ as input features and their measured latencies as ground-truth labels.

\subsection{Surrogate-based Evaluation}
\label{sec:surrogate_based_evaluation}
{
\setlength{\parindent}{0cm}
\textbf{Unsupervised Device Clustering.}
Given a set of homogeneous edge devices $\mathcal{C}$, HDAP first deploys a benchmark model on each device and records the latency $f_i(M)$, forming input features for clustering. Devices are then grouped into $K$ clusters $\{\mathcal{C}_1, \dots, \mathcal{C}_K\}$ via DBSCAN~\cite{DBSCAN2}, significantly reducing the evaluation burden.
}

{
\setlength{\parindent}{0cm}
\textbf{Supervised Learning for Surrogates.}
For each cluster $\mathcal{C}_k$, a surrogate model $g'_k(\cdot;\theta_k)$ is trained to predict the average latency $g_k(M')$ of a pruned model $M'$. To collect training data, HDAP samples multiple pruning vectors $\boldsymbol{X}$, prunes the model to obtain $M'$, and evaluates $f_i(M')$ on each device in $\mathcal{C}_k$. The average latency is then computed as:
\begin{equation}
g_k(M') = \frac{1}{|\mathcal{C}_k|} \sum_{c_i \in \mathcal{C}_k} f_i(M').
\end{equation}
These $(\boldsymbol{X}, g_k(M'))$ pairs are used to train a GBRT~\cite{GBRT2} surrogate model $g'_k(\cdot;\theta_k)$.
}

\subsection{Iterative Pruning and Fine-tuning}
\label{sec:HDAP}

HDAP employs an iterative framework to obtain the final pruned model, as illustrated in Fig.~\ref{fig:pipeline}. At each iteration $t \in \{1, \dots, T\}$, a population-based pruning strategy guided by NCS~\cite{NCS} is used to solve \eqref{eq:optimization_problem}. The best pruned model is then fine-tuned to recover accuracy.

In each iteration, the reference model $M$ is first encoded as a pruning vector $\boldsymbol{X}_1 = \{0, \dots, 0\}$. Based on $\boldsymbol{X}_1$, new candidates $\boldsymbol{X}_2, \dots, \boldsymbol{X}_n$ are generated via evolutionary operators. Each vector $\boldsymbol{X}_i$ corresponds to a pruned model $M' = P(M, \boldsymbol{X}_i)$, and its fitness is evaluated by:
\begin{equation} 
    \begin{cases} 
        \overline{f}(M'), & \text{if } \text{Acc}(M') \geq \alpha  \text{Acc}(M), \\[6pt]
        \overline{f}(M') + \dfrac{1 - \text{Acc}(M')}{1 - \alpha}, & \text{otherwise}.
    \end{cases}
\end{equation}
This fitness function penalizes candidates that fail to meet the accuracy requirement.
After evaluating all candidates, the one with the best fitness is selected as $\boldsymbol{X}^*_t$, yielding the pruned model $M^*_t = P(M, \boldsymbol{X}^*_t)$. The model is then fine-tuned and used as the reference for the next iteration.

\section{Experiment}
\label{sec:experiments}

\subsection{Setup} \label{sec:exp_setup}
{
\setlength{\parindent}{0cm}
\textbf{Datasets and DNN models.} We use ResNet50~\cite{ResNet} and MobileNetV1~\cite{MobileNetV1} on the ImageNet~\cite{ImageNet} for large-scale compression evaluations. For the ablation study, we use ResNet56, VGG16~\cite{VGG} on CIFAR-10~\cite{CIFAR10}, and YOLOv8n~\cite{yolov8} on Pascal VOC~\cite{VOC}, covering both image classification and object detection tasks.
}

{
\setlength{\parindent}{0cm}
\textbf{Homogeneous Edge Devices.} 
We use 10 NVIDIA Jetson Xavier NX and 10 NVIDIA Jetson Nano devices, all configured with identical software environments.
}

{
\setlength{\parindent}{0cm}
\textbf{HDAP Settings.} 
For HDAP, we set $T = 20$, population size $n = 10$, NCS iteration $G = 100$ and an accuracy ratio of $\alpha = 0.5$ across all experiments. During fine-tuning, we use the SGD optimizer with a momentum of 0.9, training for 90 epochs with an initial learning rate of 0.01, reduced by a factor of 10 every 30 epochs and the weight decay is $1 \times 10^{-4}$.
}

\subsection{ImageNet Compression Results}
\label{sec:imagenet_compression}

We evaluate HDAP on both ResNet50 and MobileNetV1 using the ImageNet dataset to demonstrate its generality and effectiveness across different model scales. The results, summarized in Table~\ref{tab:combined_compression}, show that HDAP consistently achieves the lowest average inference latency across all device clusters and FLOPs constraints, outperforming existing methods under each computational budget.
For ResNet50, HDAP yields significant latency reductions under three FLOPs settings: $1.24\times$ speedup at 3.0G, $1.69\times$ at 2.0G, and an impressive $2.86\times$ at 1.0G. Similarly, on MobileNetV1, HDAP achieves a latency of 42.54\,ms, surpassing MetaPruning and AMC with a $1.67\times$ speedup. These results highlight HDAP’s strength in optimizing inference efficiency, particularly for latency-sensitive applications.

\begin{table}[th]
\centering
\fontsize{10.5}{15}\selectfont
\resizebox{1.0\linewidth}{!}{
\begin{tabular}{ccccccc}
\toprule
\textbf{Method} & \textbf{FLOPs} & \textbf{Base} & \textbf{Pruned} & \textbf{Top-1} & \textbf{Latency} & \textbf{Speedup} \\
 &  & Top-1 (\%) & Top-1 (\%) & $\downarrow$(\%) & (ms) & \\
\midrule
\multicolumn{7}{c}{\textbf{ResNet50 on ImageNet}} \\
\midrule
AUTOSLIM~\cite{Autoslim}   & 3.00  & -     & 76.00 & -     & 244.14 & 1.21$\times$ \\
MetaPruning~\cite{MetaPruning} & 3.00  & 76.60 & 76.20 & -0.40 & -      & - \\
HALP~\cite{HALP}           & 3.02  & 77.20 & 77.44 & +0.24 & 248.20 & 1.19$\times$ \\
\textbf{HDAP (Ours)}       & 3.05  & 76.13 & 76.20 & +0.07 & \textbf{238.30} & \textbf{1.24$\times$} \\
\midrule

MetaPruning~\cite{MetaPruning} & 2.01  & 76.60 & 75.40 & -1.20 & 187.77 & 1.57$\times$ \\
AUTOSLIM~\cite{Autoslim}   & 2.00  & -     & 75.60 & -     & 183.97 & 1.61$\times$ \\
GReg2~\cite{Greg2}         & 1.81  & 76.13 & 75.36 & -0.77 & 204.60 & 1.45$\times$ \\
HALP~\cite{HALP}           & 1.98  & 77.20 & 76.47 & -0.73 & 176.49 & 1.68$\times$ \\
\textbf{HDAP (Ours)}       & 1.98  & 76.13 & 75.15 & -0.98 & \textbf{175.00} & \textbf{1.69$\times$} \\
\midrule

MetaPruning~\cite{MetaPruning} & 1.04  & 76.60 & 73.40 & -3.20 & 126.54 & 2.34$\times$ \\
AUTOSLIM~\cite{Autoslim}   & 1.00  & -     & 74.00 & -     & 117.32 & 2.52$\times$ \\
GReg2~\cite{Greg2}         & 1.37  & 76.13 & 73.90 & -2.23 & 184.11 & 1.61$\times$ \\
HALP~\cite{HALP}           & 1.12  & 77.20 & 74.41 & -2.79 & 106.18 & 2.79$\times$ \\
\textbf{HDAP (Ours)}       & 0.95  & 76.13 & 73.26 & -2.87 & \textbf{103.40} & \textbf{2.86$\times$} \\
\midrule
\multicolumn{7}{c}{\textbf{MobileNetV1 on ImageNet}} \\
\midrule
MetaPruning~\cite{MetaPruning} & 324   & 70.60 & 70.90 & +0.30 & 47.76  & 1.49$\times$ \\
AMC~\cite{AMC}             & 285   & 70.90 & 70.50 & -0.40 & 48.04  & 1.48$\times$ \\
\textbf{HDAP (Ours)}       & 311   & 70.90 & 70.22 & -0.68 & \textbf{42.54} & \textbf{1.67$\times$} \\
\bottomrule
\end{tabular}
}
\caption{Compression results for ResNet50 and MobileNetV1 on ImageNet. Best results are highlighted in \textbf{bold}. Methods marked with ‘-’ do not have published pruned structures, preventing direct latency evaluation.}
\label{tab:combined_compression}
\vspace{-0.2cm}

\end{table}

Despite aggressive pruning, HDAP maintains competitive accuracy on both models. For instance, on ResNet50 at 3.0G FLOPs, HDAP improves Top-1 accuracy by 0.07\% after pruning, while on MobileNetV1, it incurs a slight accuracy drop of 0.68\%, demonstrating a favorable trade-off between latency reduction and accuracy retention.

\begin{figure}
    \centering
    \includegraphics[ width=.9\linewidth]{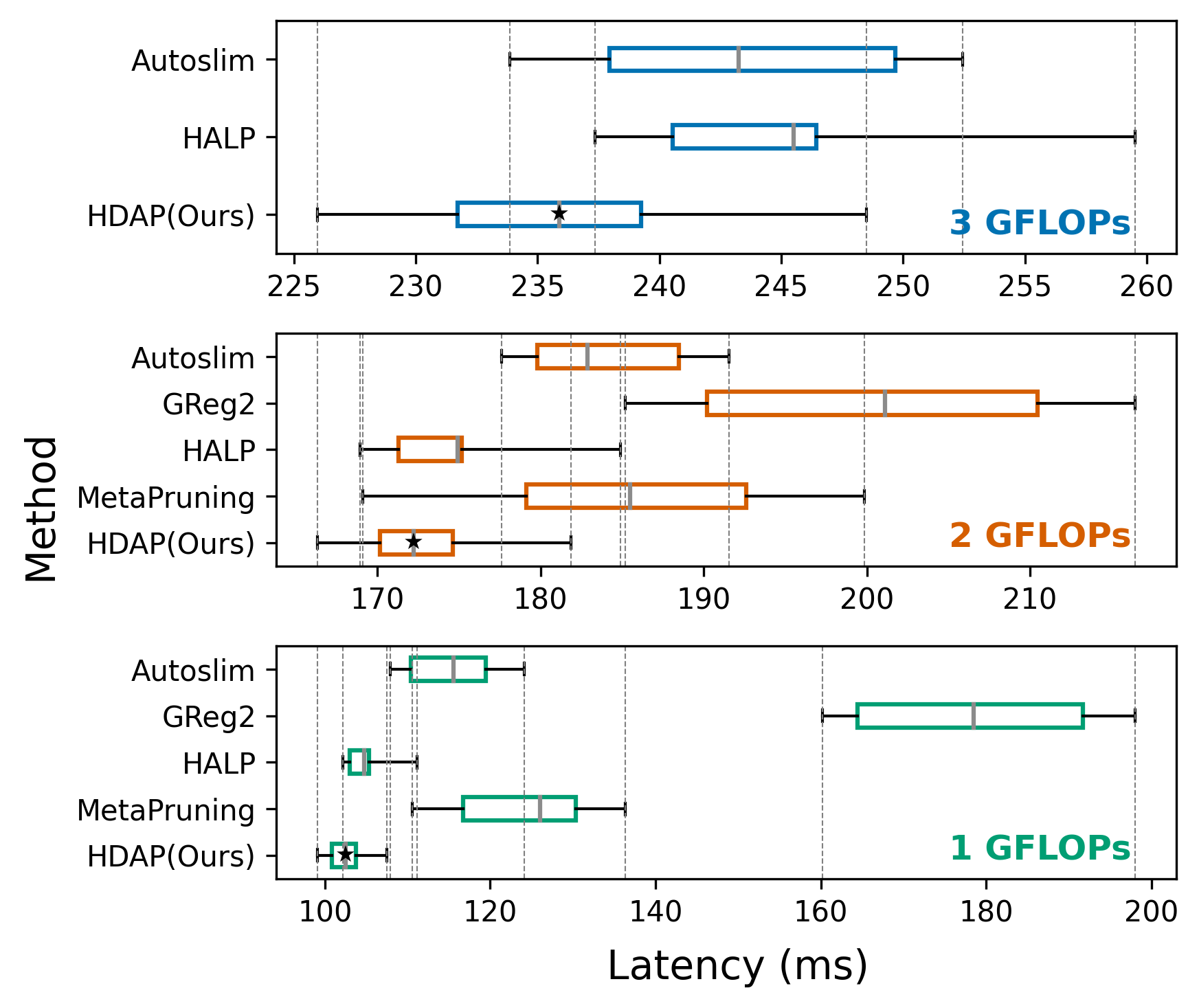}
    \vspace{-0.2cm}
    \caption{Latency distributions across device clusters for ResNet50 under three FLOPs budget. HDAP achieves the best performance, as indicated by ‘$\star$’, achieving the lowest maximum and minimum latencies across device clusters.}

    \label{fig:resnet50-cluster-result-comparison}
    \vspace{-0.3cm}
    
\end{figure}

Moreover, as illustrated in Fig.~\ref{fig:resnet50-cluster-result-comparison}, HDAP achieves the lowest maximum and minimum inference latencies among all methods across device clusters, demonstrating its robustness to hardware performance variance. This validates HDAP’s capability to provide consistently low-latency deployment on homogeneous edge devices.

\subsection{Surrogate Model Evaluation} 
\label{sec:evaluation_surrogate}

\begin{figure}
    \centering
    \includegraphics[width=.85\linewidth]{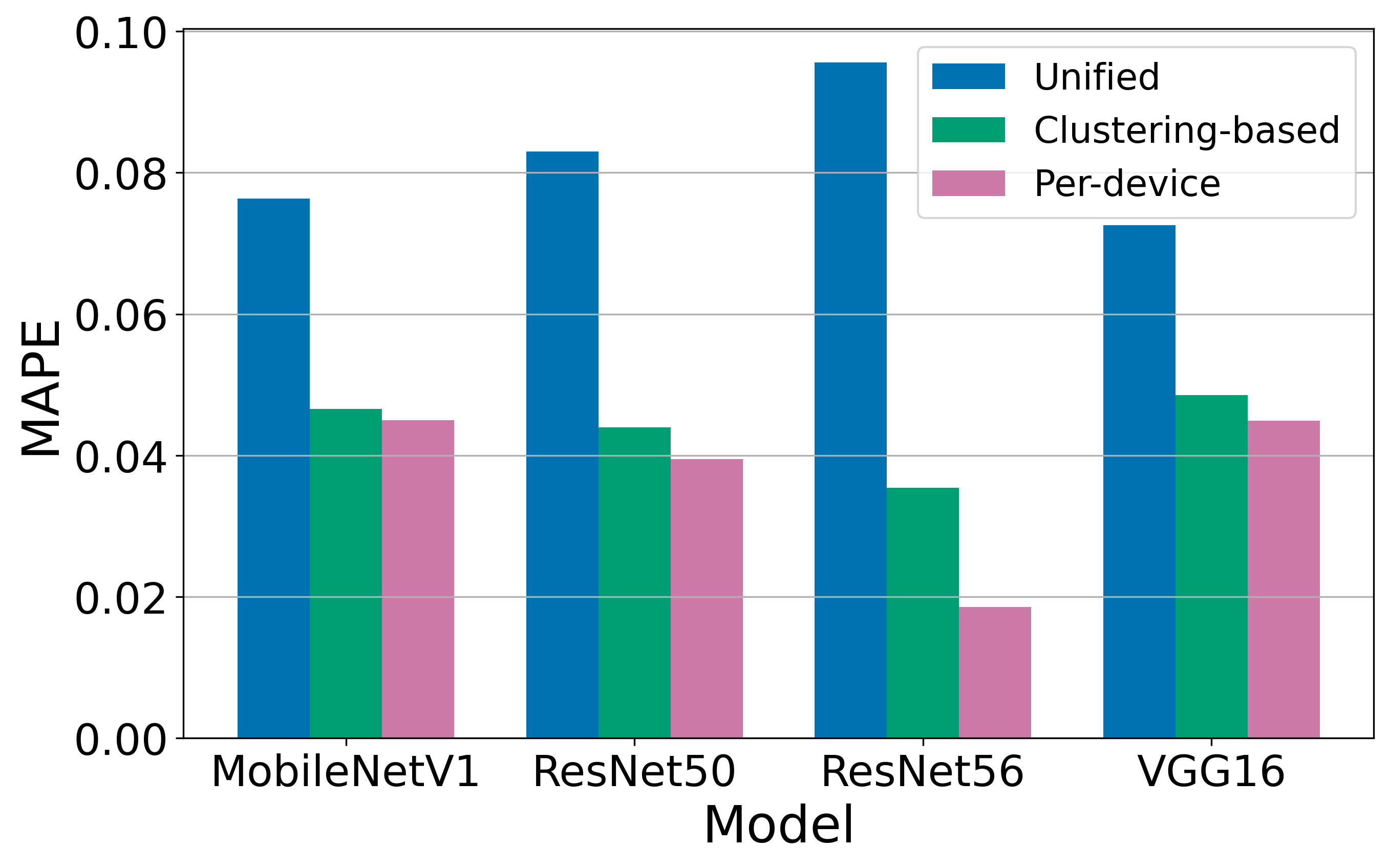}
    \vspace{-0.2cm}
    \caption{Accuracy of surrogate models for different construction methods on four DNN models. The Clustering-based method achieves prediction accuracy close to the Per-device method while outperforming the Unified method.}

    \label{fig:surrogate_methods_compare}
    \vspace{-0.4cm}
\end{figure}

We compare our Clustering-based surrogate construction with two baselines—Unified and Per-device—illustrated in Fig.~\ref{fig:surrogate_model_construction_methods}. Evaluation is conducted on four DNN models (MobileNetV1, ResNet50, ResNet56, and VGG16), using mean absolute percentage error (MAPE) between predicted and actual latency as the metric.
As shown in Fig.~\ref{fig:surrogate_methods_compare}, the Clustering-based method achieves accuracy comparable to the Per-device approach, while consistently outperforming the Unified method across all models. By grouping devices with similar latency behavior, our method better captures performance variations among homogeneous edge devices, which the Unified method fails to model.
While the Per-device method yields the lowest prediction error, it requires measurements from all devices, making it impractical in real-world deployments. In contrast, our Clustering-based approach only requires access to a representative device per cluster, offering a more scalable and deployment-friendly solution.

\subsection{Ablation Study} 
\label{sec:ablation}


We conduct an ablation study on CIFAR-10 (ResNet56, VGG16) and Pascal VOC (YOLOv8n), using grid search to compare pruning results guided by surrogate and hardware evaluation. Models undergo iterative pruning and fine-tuning for 10 iterations; CIFAR-10 models are fine-tuned for 90 epochs each iteration, while YOLOv8n is fine-tuned for 10 epochs.
As shown in Table~\ref{tab:resnet56}, surrogate evaluation closely approximates hardware-guided pruning in accuracy and latency.
For ResNet56, surrogate achieves 91-92\% accuracy with latency comparable to hardware on Jetson Xavier NX and Nano. 
VGG16 surrogate evaluation outperforms hardware accuracy (90.65\% vs. 88.76\%) with similar latency. For YOLOv8n, surrogate evaluation reaches nearly identical mAP and latency (79.50\%, 45.22 ms vs. 79.68\%, 46.29 ms).
These results confirm that surrogate evaluation effectively replaces hardware evaluation for efficient model compression.

\begin{table}[ht]

\centering
\fontsize{11}{16}\selectfont
\centering
\vspace{-0.1cm}
\resizebox{1\linewidth}{!}{
\begin{tabular}{ccccccc}
\toprule
\multicolumn{7}{c}{\textbf{ ResNet56 and VGG16 on CIFAR-10}} \\
\midrule
\multirow{2}{*}{\textbf{Model}} & \multirow{2}{*}{\textbf{Device}} & \textbf{Evaluation} & \textbf{Top-1}  & \textbf{FLOPs}  & \textbf{Latency} &  \multirow{2}{*}{\textbf{Speedup}}\\ 
 & & \textbf{Method}& \textbf{(\%)}  & \textbf{(M)} & \textbf{(ms)}   \\
\midrule
\multirow{6}{*}{ResNet56} & \multirow{3}{*}{nx} &  - & 93.53 & 125.75 & 19.93 & 1.00$\times$\\
& & Hardware & 91.66 & 44.54 & 18.37 & 1.08$\times$\\ 
& & Surrogate & 91.42 & 44.12 & 18.57 & 1.07$\times$ \\ 
& \multirow{3}{*}{nano} &  - & 93.53 & 125.75 & 44.57 & 1.00$\times$\\
& & Hardware & 92.48 & 42.68 & 38.85 & 1.14$\times$\\ 
& & Surrogate & 92.16 & 45.62 & 37.70 & 1.18$\times$ \\ 
\midrule
\multirow{3}{*}{VGG16} & \multirow{3}{*}{nx} & - & 93.97 & 398.14 & 32.82 & 1.00$\times$\\
& & Hardware & 88.76 & 7.25 & 4.95 & 6.63$\times$\\ 
& & Surrogate & 90.65 & 12.19 & 5.14 & 6.38$\times$ \\ 
\midrule
\multicolumn{7}{c}{\textbf{YOLOv8n on Pascal VOC}} \\
\midrule
\multirow{2}{*}{\textbf{Model}} & \multirow{2}{*}{\textbf{Device}} & \textbf{Evaluation} & \textbf{mAP}  & \textbf{FLOPs}  & \textbf{Latency} &  \multirow{2}{*}{\textbf{Speedup}}\\ 
 & & \textbf{Method} & \textbf{(\%)}  & \textbf{(G)} & \textbf{(ms)}   \\
\midrule
\multirow{3}{*}{YOLOv8n (2023)} & \multirow{3}{*}{nx} & - & 83.27 & 4.08 & 50.06 & 1.00$\times$\\
& & Hardware & 79.68 & 3.22 & 46.29 & 1.08$\times$\\ 
& & Surrogate & 79.50 & 3.07 & 45.22 & 1.11$\times$ \\ 
\bottomrule
\end{tabular}
}
\caption{Compression results on classification and detection tasks show that surrogate evaluation guided pruning achieves performance comparable to hardware evaluation guided methods across Jetson Xavier NX (nx) and Jetson Nano (nano).}

\label{tab:resnet56}
\vspace{-0.4cm}
\end{table}

\subsection{Acceleration of Surrogate-based Evaluation} 
\label{sec:acceleration}
\begin{figure}
    \centering
    \includegraphics[width=.9\linewidth]{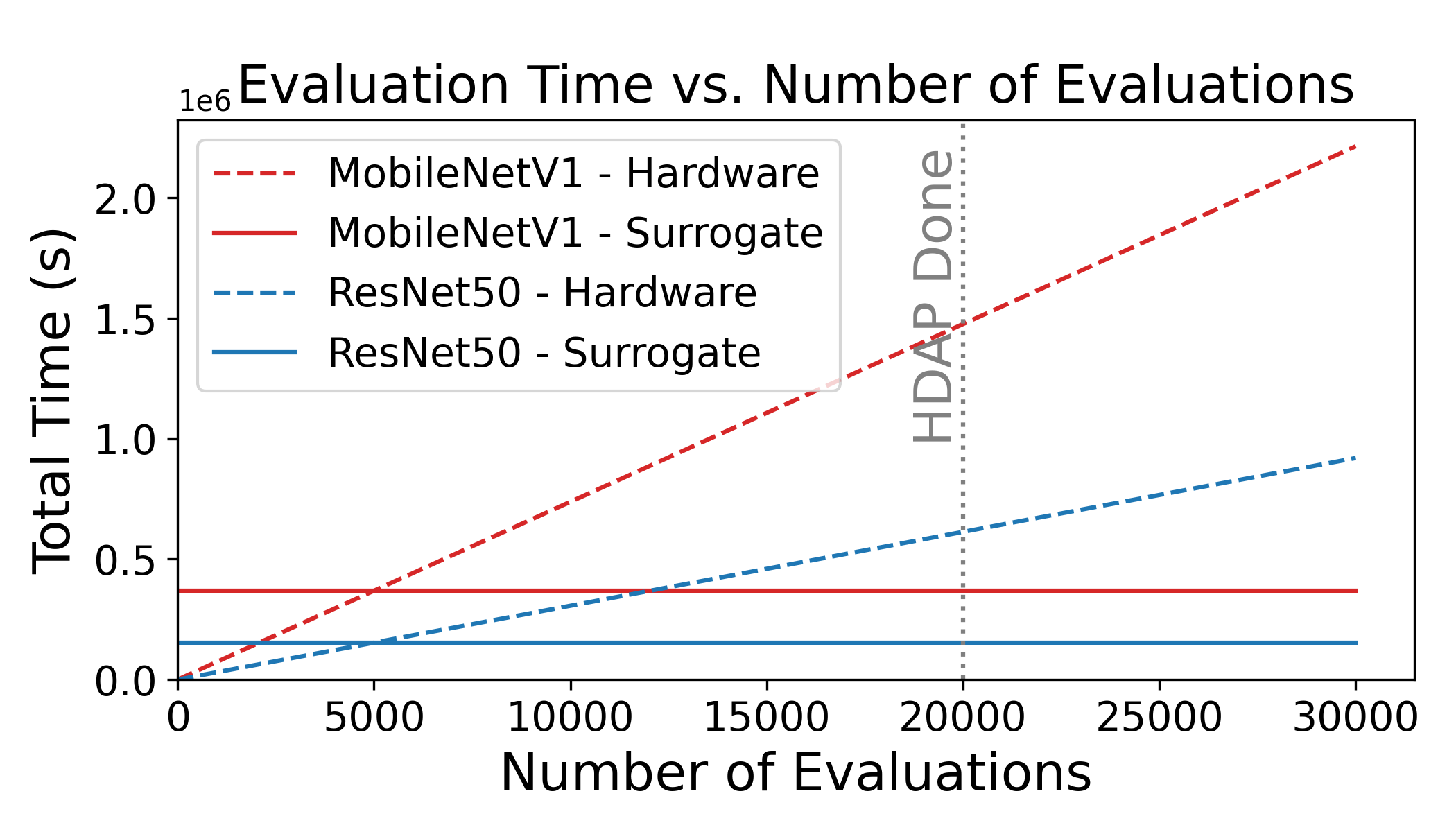}
    \vspace{-0.2cm}
    \caption{Cumulative evaluation time during HDAP: surrogate evaluation maintains constant, while hardware evaluation grows linearly.}
    \label{fig:evaluation_time_comparison}
    \vspace{-0.2cm}

\end{figure}

To evaluate efficiency, we compare per-candidate latency evaluation time between surrogate and hardware-based methods. As shown in Table~\ref{tab:acceleration}, surrogate evaluation achieves significant speedups, up to $2.65 \times 10^7$ for MobileNetV1 and $1.36 \times 10^7$ for ResNet50.
Fig.~\ref{fig:evaluation_time_comparison} shows cumulative evaluation time over the full HDAP process. The surrogate is built using 5{,}000 hardware measurements (training takes $\sim$5s) and used for 20{,}000 evaluations. While hardware cost grows linearly, surrogate evaluation remains nearly constant.
These results confirm that surrogate-based evaluation offers substantial runtime savings with negligible overhead, enabling fast and scalable pruning across homogeneous edge devices. 

\begin{table}[ht]
\centering
\fontsize{11}{16}\selectfont
\resizebox{1.0\linewidth}{!}{
\begin{tabular}{cccc}
\toprule
 \multirow{2}{*}{\textbf{Model}} & \multicolumn{2}{c}{\textbf{Evaluate Method}}  &  \multirow{2}{*}{\textbf{Acceleration}}  \\
 &Hardware (s)&Surrogate (s)& \\
\midrule

MobileNetV1 & 73.82 & 2.79 $\times 10^{-6}$ & 2.65 $\times 10^7$ \\
ResNet50 & 30.67 & 2.25 $\times 10^{-6}$  & 1.36 $\times 10^7$ \\

\bottomrule
\end{tabular}
}
\caption{Single evaluation time: surrogate vs. hardware.}
\vspace{-0.5cm}

\label{tab:acceleration}
\end{table}

\section{Conclusion}
\label{sec:conclusion}

We propose HDAP, a hardware-aware DNN compression framework that addresses performance variation across homogeneous edge devices through device clustering and surrogate-based evaluation. Extensive experiments across multiple device types (Jetson Xavier NX and Nano) and task types (image classification and object detection) demonstrate that HDAP consistently reduces average latency while maintaining competitive accuracy, enabling efficient deployment in latency-sensitive edge scenarios.

\vspace{-0.12cm}

\section*{Acknowledgment}
This work was supported in part by the National Natural Science Foundation of China under Grant 62272210 and Grant 62331014.

\bibliographystyle{IEEEtran}
\bibliography{references}

\end{document}